\documentclass{article}

\usepackage[final]{nips_2018}

\usepackage[utf8]{inputenc} 
\usepackage[T1]{fontenc}    
\usepackage{hyperref}       
\usepackage{url}            
\usepackage{booktabs}       
\usepackage{amsfonts}       
\usepackage{nicefrac}       
\usepackage{microtype}      
\usepackage{graphicx}
\usepackage{color,soul}
\usepackage{graphicx}
\usepackage{caption}
\usepackage{subcaption}
\usepackage{adjustbox}

\usepackage{amsmath}
\usepackage{adjustbox}
\usepackage{array}
\usepackage{siunitx} 
\usepackage{tabularx}
\usepackage{todonotes}
\usepackage{xcolor}
\usepackage[normalem]{ulem}
\usepackage{titlesec}
\usepackage{multirow}
\usepackage{arydshln}
\usepackage{authblk}

\hypersetup{
  pdfinfo={
    Title={The Effect of Heterogeneous Data for
Alzheimer’s Disease Detection from Speech},
    Author={Aparna Balagopalan, Jekaterina Novikova, Frank Rudzicz, Marzyeh Ghassemi},
    Subject={},
    Keywords={Alzheimer’s Disease;Multi-task data;Data augmentation;Natural Language Processing}
  }
}
\usepackage{pdfpages}

\def\JNdel#1{\bgroup\markoverwith{\textcolor{blue}{\rule[0.5ex]{2pt}{1pt}}}\ULon{#1}}

\newcolumntype{P}[1]{>{\centering\arraybackslash}p{#1}}

\pdfoutput = 1

\title{The Effect of Heterogeneous Data for\\Alzheimer's Disease Detection from Speech }

\author[1,2]{\textbf{Aparna Balagopalan}}
\author[1]{\textbf{Jekaterina Novikova}}
\author[1,2,3,4,5]{\textbf{Frank Rudzicz}}
\author[2,3]{\textbf{Marzyeh Ghassemi}}
\affil[1]{Winterlight Labs, Toronto, ON}
\affil[2]{Department of Computer Science, University of Toronto, ON}
\affil[3]{Vector Institute for Artificial Intelligence, Toronto, ON}
\affil[4]{Li Ka Shing Knowledge Institute, St Michael’s Hospital, Toronto, ON}
\affil[5]{Surgical Safety Technologies Inc., Toronto, ON}
\affil[ ]{\texttt{\{aparna,jekaterina\}@winterlightlabs.com,\{frank,marzyeh\}@cs.toronto.edu}}

\graphicspath{ {images} }
\begin{document}

\maketitle

\begin{abstract}
Speech datasets for identifying Alzheimer's disease (AD) are generally restricted to participants performing a single task, e.g. describing an image shown to them. As a result, models trained on linguistic features derived from such datasets may not be generalizable across tasks. Building on prior work demonstrating that same-task data of healthy participants helps improve AD detection on a single-task dataset of pathological speech, we augment an AD-specific dataset consisting of subjects describing a picture with multi-task healthy data. We demonstrate that normative data from multiple speech-based tasks helps improve AD detection by up to 9\%.
Visualization of decision boundaries reveals that models trained on a combination of structured picture descriptions and unstructured conversational speech have the least out-of-task error and show the most potential to generalize to multiple tasks. We analyze the impact of age of the added samples and if they affect fairness in classification.
We also provide explanations for a possible inductive bias effect across tasks using model-agnostic feature anchors. This work highlights the need for heterogeneous datasets for encoding changes in multiple facets of cognition and for developing a task-independent AD detection model.

\end{abstract}

\section{Introduction}
\vspace{-0.4em}

Alzheimer’s disease (AD) is a neurodegenerative disease affecting over 40 million people worldwide with high costs of acute and long-term care \cite{prince2016world}. Recruitment of participants with cognitive impairment has historically been a bottleneck in clinical trials \cite{watson2014obstacles}, making AD datasets relatively small. Additionally, though cognitive assessments test domains of cognition through multiple tasks, most available datasets of pathological speech are restricted to participants performing a single task.
Picture description using an image to elicit narrative discourse samples is one such task that has proved to be successful in detecting AD \cite{forbes2005detecting}. However, it is important to develop ML models of high performance that would produce results generalizable across different tasks.

Several studies have used natural language processing and machine learning to distinguish between healthy and cognitively impaired speech of participants describing a picture. Fraser {\em et al.} \cite{fraser2016linguistic} used linguistic and acoustic features to classify healthy and pathological speech transcripts 
with an accuracy of $82\%$. Similarly, Karlekar {\em et al.} \cite{karlekar2018detecting} classified utterances of speakers as AD or healthy (HC)
with an accuracy of $91.1\%$ using an enlarged, utterance-level view of transcripts of picture descriptions. In line with previous research, we use linguistic and acoustic features of speech as input to our ML model. Furthermore, we extend the model to using data from several \emph{different} tasks.

Noorian {\em et al.} \cite{noorian2017importance} demonstrated that using within-task data of healthy participants describing a picture improved AD detection performance by up to 13\%. In this paper, we evaluate if 
model performance improves with the addition of data from healthy participants, with varying ages, performing either the same or \emph{different} tasks. We find that models trained on datasets of picture description tasks augmented with conversational speech of healthy speakers learn decision boundaries that are more generalizable across activities with lower out-of-task errors. We observe a 9\% increase in AD detection performance when normative data from different tasks are utilized. We also analyze if each task provides domain-specific inductive bias for other tasks to obtain a model setting capable of detecting AD from any sample of speech using high-precision model-agnostic explanations proposed by Ribeiro {\em et al.} \cite{ribeiro2018anchors} and computation of various error metrics related to classification. 

\vspace{-0.4em}
\section{Data}
\label{data}
\vspace{-0.4em}

\begin{table}[ht!]
    \begin{adjustbox}{width=1.00\textwidth}
    \begin{tabularx}{1.2\textwidth}{l|c|c|c|l}
    \hline
      \textbf{Dataset} & \textbf{Samples}&\textbf{Subjects} & \textbf{Age} & \textbf{Tasks in Dataset}\\
         \hline
      Dementia Bank\cite{boller2005dementiabank}\emph{(DB)} & 409 (180 AD) & 210 & 45 - 90 & Picture Description\\
      \hdashline[0.5pt/1pt]
      Healthy Aging Picture Description \emph{{(HAPD)}} & 122 & \multirow{2}{*}{50} & \multirow{2}{*}{50 - 95} & Picture Description\\
      Healthy Aging Fluency \& Paragraph tasks \emph{{(HAFP)}} & 427 & & & Verbal Fluency, Reading \\
      \hdashline[0.5pt/1pt]
      Famous People\cite{anonymous2018} \emph{(FP)} & 231 & 9 & 30 - 88 & Conversational Speech\\
\hline
\end{tabularx}
    \end{adjustbox}
    \vspace{0.1cm}
\scriptsize{\caption{Speech datasets used. Note that HAPD, HAFP and FP only have samples from healthy subjects. Detailed description in App. \ref{data}.}}
\end{table}
\label{tab:datasets}

\vspace{-0.25cm}
All datasets shown in Tab.~\ref{tab:datasets} were transcribed manually by trained transcriptionists, employing the same list of annotations and protocols, with the same set of features extracted from the transcripts (see Sec.~\ref{sec:anchors_exp}). HAPD and HAFP are jointly referred to as HA.

\vspace{-0.4em}
\section{Methods}
\label{sec:anchors_exp}
\vspace{-0.4em}

\noindent {\bf Feature Extraction:}
We extract 297 linguistic features
from the transcripts and 183 acoustic features from the associated audio files, all task-independent. Linguistic features encompass syntactic features (e.g. syntactic complexity \cite{lu2010automatic}), lexical features (e.g. occurrence of production rules). Acoustic features include Mel-frequency Cepstral Coefficients (MFCCs) \&  pause-related features (e.g., mean pause duration). We also use sentiment lexical norms \cite{warriner2013norms}, local, and global coherence features \cite{brandao2017discourse}.

\noindent {\bf Feature Predicates as Anchors for Prediction:}
Given a black box classifier $f$ with interpretable input representation, Ribeiro {\em et al.} \cite{ribeiro2018anchors} define anchors $A$ as a set of input rules such that when conditions in the rule are met, humans can confidently predict the behavior of a model with high precision. Since the inputs to the classifier are engineered features with finite ranges,
we can obtain sufficient conditions for the prediction $f(x)$ in terms of interpretable feature thresholds for an unseen instance $x$ .  Anchors are found by 
maximizing the metric of coverage, defined as the probability of anchors holding true to samples in the data distribution $p$, in \cite{ribeiro2018anchors}. Hence, $cov(A)=\mathop{\mathbb{E}}_{p(z)}[A(z)]$ is maximized, where $precision(A)>\tau$. 

We show in Sec. \ref{sec:attention} that anchors identified from a model trained on multiple tasks have more coverage over the data distribution than those obtained from a model trained on a single task. Such a scenario is possible when task-independant, clinically relevant speech features 
are selected as anchors (e.g., fraction of filled pauses in speech \cite{lee2011speech}, acoustic features \cite{rudzicz2014automatically} etc. ). Additionally, such selected anchors must also be associated with thresholds applicable across multiple types of speech.   

\vspace{-0.4em}
\section{Experiments}
\label{experiments}
\vspace{-0.4em}

Binary classification of each speech transcript as AD or HC is performed. We do 5-fold cross-validation, stratified by subject so that each subject's samples do not occur in both training and testing sets in each fold. The minority class is oversampled in the training set using 
SMOTE~\cite{chawla2002smote} 
to deal with the class imbalance. We consider a Random Forest (100 trees), Na\"ive Bayes (with equal priors), 
SVM (with RBF kernel), and a 2-layer neural network (10 units, Adam optimizer, 500 epochs)\cite{pedregosa2011scikit}.
Additionally, we augment the DB data with healthy samples from FP with varied ages.

\vspace{-0.4em}
\section{Results and Discussion}
\label{results}
\vspace{-0.4em}


\subsection{Visualization of Class Boundaries}
\vspace{-0.4em}

Since data of different tasks have different noise patterns, the probability of overfitting to noise 
is reduced with samples from different tasks. This can also be visualized as decision boundaries of models trained on various dataset combinations. For Fig.\ref{fig:decision-boundaries}, we embed the 480-dimensional feature vector into 2 dimensions using Locally Linear Embeddings~\cite{wang2012locally} trained on DB.
\vspace{-0.4em}

\begin{figure}[ht!]
    \centering
    \begin{subfigure}[t]{0.3\textwidth}
        \centering
        \includegraphics[height=1.7in,width=1.9in]{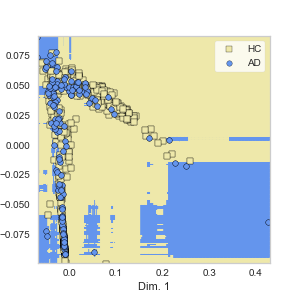}
        \caption{Pic. descriptions (PD); \textbf{28.6\%}}
        \label{fig:picture_descr}
    \end{subfigure}%
    ~ 
    \centering
    \begin{subfigure}[t]{0.3\textwidth}
        \centering
        \includegraphics[height=1.7in,width=1.9in]{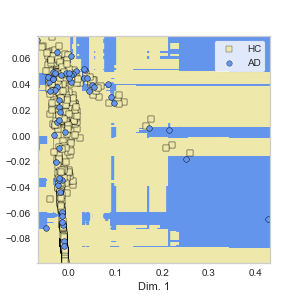}
        \caption{PD + structured tasks; \textbf{17.8\%}}
    \label{fig:fluency}
    \end{subfigure}%
 ~
    \begin{subfigure}[t]{0.3\textwidth}
        \centering
        \includegraphics[height=1.7in,width=2in]{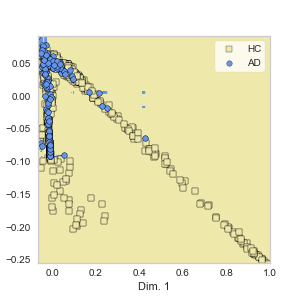}
        \caption{PD + general speech; \textbf{3.6\%}}
        \label{fig:general}
        \end{subfigure}
    \scriptsize{\caption{Decision boundaries with RF classifier for datasets with their out-of-task error shown in bold; scattered points shown belong to the train set in each case. For models trained using general, task-independent features on picture description (Fig.\ref{fig:picture_descr}) \& other structured tasks from HAFP such as fluency (Fig.\ref{fig:fluency}), decision boundaries are patchy as a result of few, far-lying points from the classes (e.g, in the fourth quadrant), leading to misclassifications on other tasks with varying feature ranges. However, on datasets consisting of general, unstructured conversations, this does not happen Fig.\ref{fig:general}.}}
\end{figure}
   \label{fig:decision-boundaries}

\vspace{-0.25cm}

In datasets consisting of picture descriptions and conversational speech (DB + FP), the feature ranges increase as compared to picture description tasks, so it is expected that a classifier trained on structured tasks only (DB + HAFP) would incorrectly classify healthy samples in the fourth quadrant (error rates for tasks not in dataset is 17.8\%). However, decision boundaries for models trained on a mix of structured tasks and unstructured conversational speech seem to be more generalizable across tasks. E.g., decision boundaries obtained from DB + FP could apply to most datapoints in HAFP (out of task error rate is 3.6\%). Clinically, some of the features used such as the patterns in usage of function words like pronouns have shown to reflect stress-related changes in gene expression, possibly caused due to dementia \cite{mehl2017natural} which would not depend on the task type and could explain such a common underlying structure to features.

\vspace{-0.4em}
\subsection{Classification Performance}
\vspace{-0.4em}

Results of binary classification with different dataset combinations (i.e., the proportion of each dataset used) are in Tab.~\ref{tab:table1}. The highest F1 score on DB is $80.5\%$ with SVM as obtained by Noorian {\em et al.} \cite{noorian2017importance}, enabling similar comparisons.

\vspace{-0.8em}

\begin{table}[!h]

\begin{center}
\begin{adjustbox}{max width=0.8\textwidth}

\vspace{0.1cm}
\small{
\begin{tabularx}{0.95\textwidth}{c c|c|c|c|c}
\hline
    
\textbf{Data} & \textbf{Size} & \textbf{Na\"ive Bayes}& \textbf{SVM}& \textbf{RF}& \textbf{NN}\\
\hline

& & F1 \hspace{0.5cm} F1&F1 \hspace{0.5cm} F1 &F1 \hspace{0.5cm} F1 &F1 \hspace{0.5cm} F1   \\ 
& & \small{(mi.)} \hspace{0.05cm} \small{(ma.)}& \small{(mi.)} \hspace{0.05cm} \small{(ma.)}& \small{(mi.)} \hspace{0.05cm} \small{(ma.)}& \small{(mi.)} \hspace{0.05cm} \small{(ma.)}\\
\hline \hline

DB & 229 HC & 62.90 \hspace{0.05cm} 60.01& \textbf{80.52 \hspace{0.05cm} 73.01}& 73.52 \hspace{0.05cm} 67.82& 75.92 \hspace{0.05cm} 72.76\\
\hdashline[0.5pt/1pt]

DB + HAPD & 351 HC & 63.99 \hspace{0.05cm} 63.37& \textbf{82.97 \hspace{0.05cm} 78.65}& 79.00 \hspace{0.05cm} 74.76& 81.39 \hspace{0.05cm} 78.07\\
DB + 0.29*HAFP & 352 HC & 60.00 \hspace{0.05cm} 58.62& 82.53 \hspace{0.05cm} 77.26& 78.94 \hspace{0.05cm} 72.80& 82.24 \hspace{0.05cm} 78.54\\
\hdashline[0.5pt/1pt]

DB + 0.42*HA & 460 HC & 65.06 \hspace{0.05cm} 63.90& 82.74 \hspace{0.05cm} 78.58& 78.09 \hspace{0.05cm} 73.70& \textbf{84.69 \hspace{0.05cm} 80.02}\\
DB + FP & 460 HC & 56.05 \hspace{0.05cm} 52.38& 83.71 \hspace{0.05cm} 80.21& 77.41 \hspace{0.05cm} 73.92& 82.19 \hspace{0.05cm} 79.26\\
\hdashline[0.5pt/1pt]

DB + HAFP & 656 HC & 74.08 \hspace{0.05cm} 70.97& \textbf{87.43 \hspace{0.05cm} 80.21}& 82.59 \hspace{0.05cm} 77.60& 86.53 \hspace{0.05cm} 79.53\\
\hdashline[0.5pt/1pt]

DB + HA & 778 HC & 74.63 \hspace{0.05cm} 70.33& \textbf{89.31 \hspace{0.05cm} 82.32}& 85.62 \hspace{0.05cm} 76.44& 88.08 \hspace{0.05cm} 80.50\\
\hline

\end{tabularx}}
\end{adjustbox}
\vspace{0.8em}

\scriptsize{\caption{AD vs HC classification. Highest F1 scores are shown in bold for datasets of similar size.}}

\end{center}
\end{table}
\label{tab:table1}

\vspace{-0.8em}

We see the same trend of increasing model performance with normative data from the picture description task, as shown by Noorian {\em et al.} \cite{noorian2017importance}. We observe that this increase is independent of the nature of the task performed -- normative picture description task data of similar size as in \cite{noorian2017importance} and the same amount of normative data from different structured tasks of fluency tests and paragraph reading prove to be helpful, bringing about a similar increase in scores (+2\%, +5\% absolute F1 micro and macro)\footnote{Friedman chi-squared = 2, df = 1, p-value = 0.1573; not significant difference}. Interestingly, performance of detecting the majority (healthy) class (reflected in F1 micro) as well as the minority (AD) class (reflected in F1 macro) increases with additional data.

Augmenting DB with same amount of samples from structured tasks (HA) and from conversational speech (FP) brings about similar performance\footnote{Friedman chi-squared = 0, df = 1, p-value = 1; not significant difference}. Doubling the initial amount of control data with data from a different structured task (HA, HAFP) results in an increase of up to 9\% in F1 scores. 

\vspace{-0.4em}
\subsection{Impact of Age}
\vspace{-0.6em}

\begin{table}[ht!]
\vspace{-0.4em}
\scriptsize{}
     \begin{center}
 \begin{adjustbox}{max width=0.48\textwidth}
    \begin{tabularx}{0.48\textwidth}{c|c|c|c}
    \hline
      \textbf{Age bin} & \textbf{Number of samples}&\textbf{F1(mi.)} & \textbf{F1(ma.)}\\
      \hline
      \hline
      30 - 45 years & 50 & 72.37 & 68.36\\
      45 - 60 years & 50 & 77.20 & 74.49\\
      60 - 75 years & 50 & 81.45 & 79.47\\
      75 - 90 years & 50 & 81.48 & 79.38\\
\hline
    \end{tabularx}
\end{adjustbox}
  \end{center}
  \scriptsize{\caption{Augmenting DB with healthy data of varied ages. Scores averaged across 4 classifiers.}}
\end{table}
\label{tab:age}

\vspace{-0.4em}

We augment DB with healthy samples from FP with varying ages (Tab.\ref{tab:age}), considering 50 samples for each 15 year duration starting from age 30.
Adding the same number of samples from bins of age greater than 60 leads to greater increase in performance. This could be because the average age of participants in the datasets (DB, HA etc.) we use are greater than 60. Note that despite such a trend, addition of healthy data produces fair classifiers with respect to samples with age$<$60 and those with age$>$60 (balanced F1 scores of 75.6\% and 76.1\% respectively; further details in App. \ref{fairness}.)



\vspace{-0.4em}
\subsection{Inductive Bias of Tasks}
\vspace{-0.4em}

Each task performed in the datasets is designed to assess different cognitive functions, e.g. fluency task is used to evaluate the ability to organize and plan \cite{whiteside2016verbal}
and picture description task -- for detecting discourse-related impairments \cite{ghayoumi2015persuasive}. As a result, it is expected that the nature of decision functions and feature predicates learned on data of each of these tasks would be different. Performance of AD identification with addition of normative data from multiple tasks (Tab.~\ref{tab:table1}), despite the possibly different nature of decision functions, suggests that training the model with samples from each task provides domain-specific inductive bias for other tasks. 
We study possible underlying mechanisms responsible for this, suggested by Caruana {\em et al.}\cite{caruna1993multitask} and Ruder {\em et al.} \cite{ruder2017overview}.

\noindent {\bf Attention-focusing on Relevant Features:}
\label{sec:attention}
Ruder {\em et al.}\cite{ruder2017overview} claim that in a small, high-dimensional dataset, information regarding relevance or irrelevance of particular features is difficult to capture. However, data related to multiple tasks can help identify features relevant across different activities. We can use anchor variables~\cite{ribeiro2018anchors} to show this effect. The coverage of features anchoring the prediction of an instance indicates the applicability of the feature predicate to the rest of the data distribution and hence the importance of the feature across the data distribution. The coverage of the anchors selected for a test set which is 10\% (50 samples) of DB changes by 40.8\% (from 0.05 to 0.07) on the addition of the HA, which indicates that there is an attention focusing effect. 

\noindent {\bf Representation bias:}
As shown by Schulz {\em et al.}\cite{schulz2018multi}, models trained on data from multiple tasks perform better than with single-task information when little training data is available for the main task. The non-linear trend of increase in model performance with the addition of different amounts of data is shown in App.\ref{fig:data_req}. 
The F1 micro score of the best performing model trained on DB + HA is 82.28\% for picture description tasks, 95.4\% for paragraph reading and 97.01\% for fluency tasks. This shows greater than trivial performance for each task and improvement in performance for picture description task from training a model purely on DB. 
Such an effect helps the model achieve non-trivial performance on AD detection for novel tasks measuring multiple domains of cognition, given a sufficiently large number of training tasks according to algorithms provided by Baxter {\em et al.}\cite{baxter2000model}. 
Hence, training models on many speech-based tasks could help develop an algorithm capable of detecting AD from any sample of spontaneous speech.

Ongoing work is on detailed analysis of nature and polarity of feature trends across various speech tasks. 
Future work will focus on learning interpretable latent representations based on the observations made, capable of good predictive performance across a multitude of tasks.

\bibliographystyle{plain}
\bibliography{bib}

\newpage
\appendix
\section{Appendix}

\subsection{Detailed Description of Datasets}
\noindent {\bf DementiaBank (DB):}
The DementiaBank\footnote{https://dementia.talkbank.org} dataset is the largest available public dataset of speech for assessing cognitive impairments. It consists of narrative picture descriptions from participants aged between 45 to 90 \cite{becker1994natural}. In each sample, a participant describes the picture that they are shown. Out of the 210 participants in the study, 117  were diagnosed with AD ($N = 180$ samples of speech) and 93 were healthy (HC; $N = 229$ samples) with many subjects repeating the task with an interval of a year. Demographics of age, sex, and years of education are provided in the dataset.  

\noindent {\bf Healthy Aging (HA) :}
The Healthy Aging dataset consists of speech samples of cognitively healthy participants ($N=50$) older than 50 years. Each participant performs three structured tasks -- picture description (HAPD), verbal fluency test\footnote{During the fluency test, the experimenter asks the participant to say aloud as many names of items belonging to a certain category and as many words as possible starting with a specific letter in a 1 minute trial}, and a paragraph reading task. Fluency and paragraph tasks are jointly referred to as HAFP.
The average number of samples per participant is 14.46. The dataset constitutes 8.5 hours of total audio.

\noindent {\bf Famous People (FP):}
The Famous People dataset \cite{anonymous2018} consists of publicly available spontaneous speech samples from 9 famous individuals (e.g., Woody Allen \& Clint Eastwood) over the period from 1956 to 2017, spanning periods from early adulthood to older age, with an average of 25 samples per person. We use speech samples of these subjects who are considered to be healthy ($N=231$), given an absence of any reported diagnosis or subjective memory complaints. This healthy control (HC) group covers a  variety of speaker ages, from 30 to 88 ($\mu = 60.9$, $\sigma = 15.4$).

\subsection{Features}
A list of 480 features belonging to three groups - acoustic, semantic/ syntactic and lexical. These features include constituency-parsing based features, syntactic complexity features extracted using Lu Syntactic Complexity analyzer \cite{lu2010automatic}, MFCC means, variances and other higher order moments.
Few of these features are listed below :
\begin{itemize}
\item Phonation rate : Percentage of recording that is voiced.
\item Mean pause duration : Mean duration of pauses in seconds.
\item Pause word ratio : Ratio of silent segments to voiced segments.
\item Short pause count normalized : Normalized number of pauses less than 1 second.
\item Medium pause count normalized : Normalized number of pauses between 1 second and 2 seconds in length.
\item ZCR kurtosis : Kurtosis of Zero Crossing Rate (ZCR) of all voiced segments across frames.
\item MFCC means : Mean of velocity of MFCC coefficient over all frames (this is calculated for multiple coefficients).
\item MFCC kurtosis: Kurtosis of mean features.
\item MFCC variance: Variance of acceleration of frame energy over all frames.
\item Moving-average type-token ratio (MATTR): Moving average TTR (type-token ratio) over a window of 10 tokens.
\item Cosine cutoff : Fraction of pairs of utterances with cosine distance $\leq$ 0.001.
\item Pauses of type `uh' : The number of `uh' fillers over all tokens.
\item Numbers of interjections/numerals : The number of interjections/numerals used over all tokens.
\item Noun ratio: Ratio of number of nouns to number of nouns + verbs.
\item Temporal cohesion feature : Average number of switches in tense.
\item Speech graph features : Features extracted from graph of spoken words in a sample including average total degree, number of edges, average shortest path, graph diameter (undirected) and graph density.
\item Filled pauses : Number of non-silent pauses.
\item Noun frequency : Average frequency norm for all nouns.
\item Noun imageability: Average imageability norm for all nouns.
\item Features from parse-tree : Number of times production rules such as number of noun phrases to determiners occurrences, occur over the total number of productions in the transcript's parse tree.
\item Syntactic complexity features: Ratio of clauses to T-units\footnote{T-unit: Shortest grammatically allowable sentences into which writing can be split or minimally terminable unit}, Ratio of clauses to sentences etc.~\cite{lu2010automatic}
\end{itemize}

\subsection{Hyper-parameters:}
Gaussian Naive Bayes with balanced priors is used.

The random forest classifier fits 100 decision trees with other default parameters in \cite{pedregosa2011scikit}.

SVM is trained with radial basis function kernel, regularization parameter $C=1$ and $\gamma = 0.001$.

The NN consists of one hidden layer of 10 units. The \emph{tanh}
activation function is used at each hidden layer. The network
is trained using Adam for 100 epochs with other default parameters in \cite{pedregosa2011scikit}.

\subsection{Effect of data from different tasks:}
\label{fig:data_req}

The effect of augmenting DB with data from a different structured task (HAFP) is shown in \ref{fig:data_req}.
\begin{figure}[ht!]
\centering
    \includegraphics[width=3.0in]{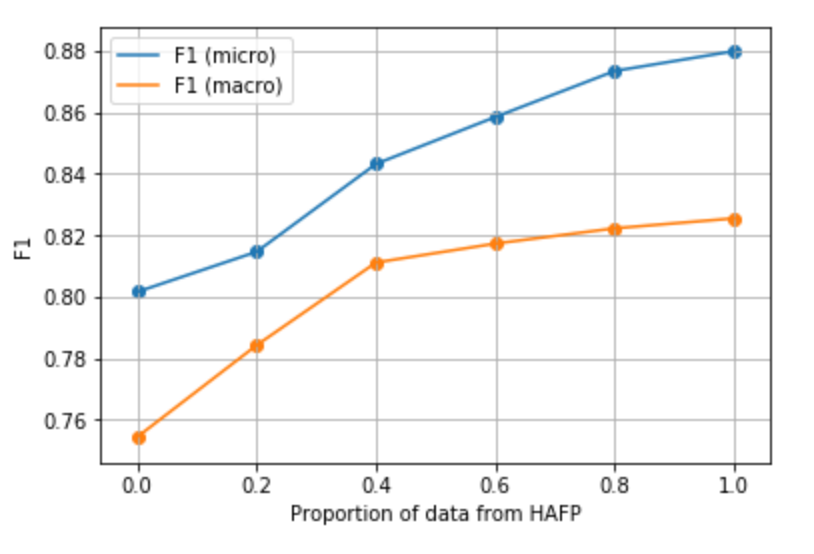}
    \captionsetup{labelformat=empty}
    \caption{Figure A.4: Effect of addition of data from a different structured task on F1 (micro) and F1 (macro)}
\end{figure}
\newline F1 scores (micro and macro) increase non-linearly with the addition of data.

\subsection{Fairness with Respect to Age:}
\label{fairness}
We evaluate fairness of classification with respect to two groups - samples with age$<$60 and those with age$>$60. A fair classifier would produce comparable classification scores for both groups. 
For the best performing classifier on DB, the F1 (micro) score for samples with age$<$60 is 85.9\% and with age$>$60 is 76.4\%. With the addition of HA, the F1 (micro) score for samples with age$<$60 and with age $>60$ is more balanced (75.6\%, 76.1\% respectively) for the same set of data points from DB. Note that the average age in both datasets are similar ($\mu_{age} \approx 68$).

\end{document}